\documentclass[runningheads]{llncs}
\usepackage[T1]{fontenc}
\usepackage{graphicx,verbatim}
\usepackage{tabularray}
\usepackage{multirow}

\renewcommand{\thefootnote}{\fnsymbol{footnote}}  

\begin{document}
\title{PathCoT: Chain-of-Thought Prompting for Zero-shot Pathology Visual Reasoning}
%

\author{Junjie Zhou\inst{1,2} 
\and Yingli Zuo\inst{1,2} 
\and Shichang Feng\inst{1,2} 
\and Peng Wan\inst{1,2} 
\and Qi Zhu\inst{1,2} 
\and Daoqiang Zhang\inst{1,2}\textsuperscript{*} 
\and Wei Shao\inst{1,2}\textsuperscript{*}} 

\authorrunning{JJ Zhou et al.}
%
\institute{The College of Artificial Intelligence, Nanjing University of Aeronautics and Astronautics \and The Key Laboratory of Brain-Machine Intelligence Technology, Ministry of Education}
\renewcommand{\thefootnote}{*}
\footnotetext[1]{Corresponding authors.}
\renewcommand{\thefootnote}{\arabic{footnote}} 

\maketitle              
\begin{abstract}
With the development of generative artificial intelligence and instruction tuning techniques, multimodal large language models (MLLMs) have made impressive progress on general reasoning tasks. Benefiting from the chain-of-thought (CoT) methodology, MLLMs can solve the visual reasoning problem step-by-step. However, existing MLLMs still face significant challenges when applied to pathology visual reasoning tasks: 
(1) LLMs often underperforms because they lack domain-specific information, which can lead to model hallucinations.
(2) The additional reasoning steps in CoT may introduce errors, leading to the divergence of answers.
To address these limitations, we propose PathCoT, a novel zero-shot CoT prompting method which integrates the pathology expert-knowledge into the reasoning process of MLLMs and incorporates self-evaluation to mitigate divergence of answers. 
Specifically, PathCoT guides the MLLM with prior knowledge to perform as pathology experts, and provides  comprehensive analysis of the image with their domain-specific knowledge. By incorporating the experts' knowledge, PathCoT can obtain the answers with CoT reasoning. Furthermore, PathCoT incorporates a self-evaluation step that assesses both the results generated directly by MLLMs and those derived through CoT, finally determining the reliable answer. 
The experimental results on the PathMMU dataset demonstrate the effectiveness of our method on pathology visual understanding and reasoning.


\end{abstract}
\section{Introduction}
In recent years, the development of generative artificial intelligence has led to the emergence of multimodal large language models (MLLMs), which are capable of processing multimodal inputs, such as images and text, by aligning visual representations with the input space of large language models (LLMs) \cite{li2023blip,liu2024visual,dai2023instructblip,zhu2023minigpt}.
Facilitated by instruction tuning, models like LLaVa \cite{liu2024visual} and GPT-4V \cite{achiam2023gpt} have showcased advanced capabilities in human interaction and reasoning, which enables MLLMs to follow human instructions and complete real-world tasks. 
These advancements have fueled enthusiasm within the AI community, leading to the emergence of MLLMs in medical fields, including radiology \cite{guo2024llava,lee2025cxr} and pathology \cite{sun2024pathasst,seyfioglu2024quilt,lu2024multimodal}, where they hold promise for advancing diagnostic support and complex reasoning.

Chain-of-Thought (CoT) methods have emerged as a powerful tool to enhance the reasoning capabilities of LLMs, achieving notable success in natural language processing (NLP) \cite{kojima2022large,wei2022chain,zhang2022automatic}. It follows a divide-and-conquer approach, breaking down complex problems into multi-step reasoning tasks. Recently, several studies have extended CoT to MLLMs, significantly boosting their reasoning abilities \cite{zhang2023multimodal,yang2023good,zheng2023ddcot,zhang2024cocot}.
For instance, MM-CoT \cite{zhang2023multimodal} first proposes a two-stage framework on complex reasoning tasks with language and vision modalities by separating rationale generation and answer inference. 
CCoT \cite{mitra2024compositional} generates the scene graph (SG) to describe the visual scene, and then integrates it as the prompt to produce a response for multimodal and compositional visual reasoning.
CoCoT \cite{zhang2024cocot} focuses on multiple image inputs that can prompt LMMs to discern and articulate the similarities and differences among inputs, laying the groundwork for answering multi-image-based questions.

Although much progress has been achieved for general visual reasoning tasks, MLLMs still face significant challenges when applied to pathology visual reasoning tasks:

1. Existing MLLMs underperform on pathology visual reasoning tasks due to the lack of domain-specific knowledge. For instance, the recent work \cite{sun2024pathmmu} curates a multimodal dataset PathMMU for understanding and reasoning in pathology, and find it that current MLLMs still struggle with the PathMMU. Even the most advanced models exhibit a significant performance gap compared to professional pathologists \cite{sun2024pathmmu}. 
This is primarily because current LLMs are trained on general-domain data that may generate incorrect answers or even hallucinations when working on specialized pathology reasoning tasks.
One potential solution is to fine-tune models on specialized pathology datasets. However, the traditional fine-tuning paradigm not only requires the collection of domain-specific data but also consumes substantial computational resources, which has limitations in scalability.

2. The additional reasoning steps in CoT may introduce errors, leading to divergence of answers. We find that the intermediate reasoning steps in CoT reasoning may introduce additional incorrect knowledge due to model hallucinations (see Fig.~\ref{fig: cases}). This causes the divergence of results between the answers derived by CoT and those from direct reasoning, and in some cases, the CoT-based results may even be worse than the direct reasoning results (see Tab.~\ref{tab: comparisons}). This divergence of answers in pathology reasoning can lead to serious consequences, such as misdiagnoses, inappropriate treatments, or delays in critical interventions \cite{xia2025cares}. Consequently, it is essential to re-evaluate CoT-based results for assessing their reliability.

To address the above limitations, we propose PathCoT, a novel CoT prompting method for zero-shot pathology visual reasoning without any fine-tuning. Specifically, PathCoT firstly generates the pathology image caption which incorporates question-agnostic and question-dependent descriptions. For better analysis of pathology images, PathCoT introduces four pathology experts: Cellular Expert, Tissue Expert, Organ Expert and Biomarker Expert. These experts can provide comprehensive analysis of pathology images with their domain-specific knowledge. By the combination of image caption and experts knowledge, PathCoT can derive the answer with CoT reasoning.  
Finally, PathCoT performs a self-evaluation process that considers both the results directly generated by MLLMs and those derived through CoT for the determination of the final answer.
We conduct experiments on the PathMMU dataset \cite{sun2024pathmmu}, which is a massive multimodal expert-level benchmark for understanding and reasoning in pathology. The experimental results demonstrate the effectiveness of our method on pathology visual understanding and reasoning.

\begin{figure}[tbp]
  \centering
  \includegraphics[width=1.\linewidth]{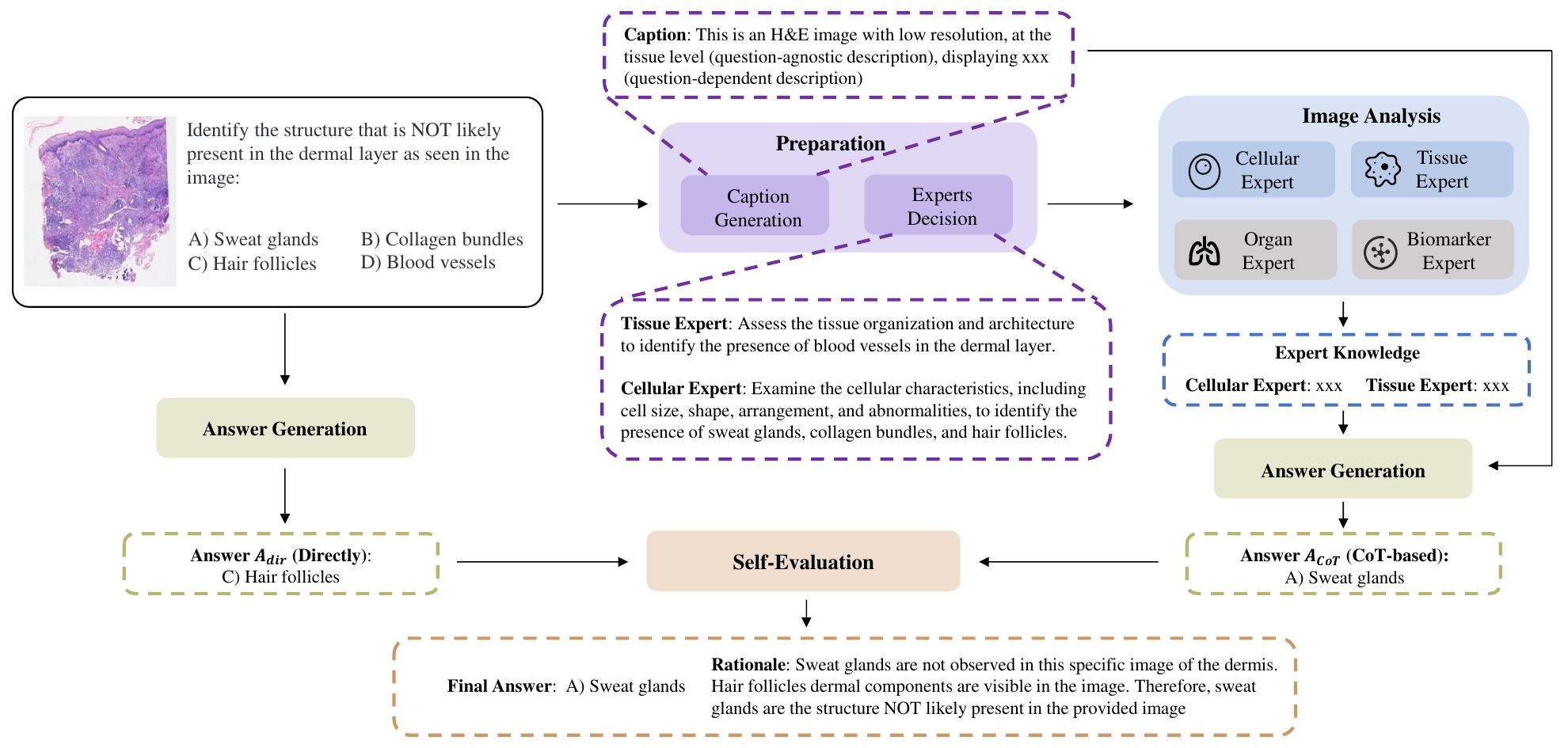}
   \caption{The framework of our proposed PathCoT.}
   \label{fig: framework}
\end{figure}

\section{Method}
\subsection{Overview}
Given the question $Q$ and image $I$, PathCoT aims to generate their corresponding answer. The input of PathCoT consists of two modalities: text and image. In addition to the question $Q$, the text modality also includes supplementary information, such as captions, expert-knowledge, etc. 
The overall framework of PathCoT is shown in Fig.~\ref{fig: framework}. The framework consists of four stages: Preparation (Sec.~\ref{sec: Preparation}), Image Analysis (Sec.~\ref{sec: Analysis}), Summary \& Answer Generation (Sec.~\ref{sec: summary}) and Self-Evaluation (Sec.~\ref{sec: Self-Evaluation}).

\subsection{Preparation}
\label{sec: Preparation}
PathCoT initially prepares for CoT reasoning, consisting of two tasks: caption generation and expert decision. Specifically, the caption generation task identifies descriptive pathological features, while the expert decision task determines the appropriate experts for subsequent image analysis.
Previous studies \cite{yang2023good,wei2024mc} demonstrate that the descriptive information can help MLLMs better understand the image and accurately answer the question. Therefore, PathCoT generates the image caption, denoted as $D_{cap}$, which contains descriptive information of the pathology image.
Particularly, we consider two types of descriptions in $D_{cap}$: question-agnostic description $D_{qa}$ and question-dependent description $D_{qd}$. Thus, the image caption $D_{cap}$ can be formulated as: $D_{cap}=\{D_{qa}, D_{qd}\}$. 

Question-agnostic description $D_{qa}$ is independent of the question $Q$, and contains basic information about the pathology image, such as image type, field of view, etc. Specifically, we can prompt the MLLMs with the following instructions: \emph{Identify the type of pathology image (e.g., HE, IHC, Gross Pathology, etc), assess the image quality (e.g., resolution, clarity, and any noise), and specify the scope of observation (cellular, tissue, or organ level).}

On the other hand, the question-dependent description $D_{qd}$ focuses on identifying pathological image features that may be relevant to the given question $Q$. For instance, if the question is related to the presence of tumor cells in the image, the question-dependent description would focus on identifying abnormal cell morphology, mitotic activity, and staining patterns indicative of malignancy. The following prompt can be used to generate $D_{qd}$: \emph{Describe the key features of the image that are relevant to the question. Initial observations should focus on identifying the primary components, staining patterns, and any notable abnormalities visible within the provided field.}

Additionally, PathCoT prepares for determining appropriate experts based on the given question for subsequent analysis. 
Specifically, PathCoT firstly identifies each expert and its corresponding responsibility (detailed in Sec.~\ref{sec: Analysis}). 
Then, by taking the image, question and experts' responsibilities as input, PathCoT generates the decision information, denoted as $I_{dec}$. This decision information contains the selection and guidance results, where the selection results indicate whether the given question can be addressed using the knowledge of the corresponding expert and the guidance $I_{G}$ provides instructions on how to leverage the selected experts to analyze the image. This guidance is subsequently used to direct the MLLM in generating the corresponding expert-knowledge.

\subsection{Image Analysis}
\label{sec: Analysis}
Pathology visual reasoning requires a comprehensive analysis of pathology images, encompassing various biological levels—from individual cells to entire organs, as well as the mechanisms related to biomarkers. To effectively address pathology-related visual reasoning tasks and enhance the analysis of pathology images, PathCoT introduces the following four experts: \textbf{Cellular Expert} analyzes cellular characteristics, including cell size, shape, arrangement, and abnormalities. 
\textbf{Tissue Expert} focuses on the organization and architecture of tissue, examining the spatial relationships between components like epithelium, stroma, and blood vessels. 
\textbf{Organ Expert} specializes in the analysis of organ-level structures, assessing the overall anatomical integrity and functional zones of the organ.
\textbf{Biomarker Expert} is responsible for identifying  crucial biomarkers for diagnosis, prognosis, and understanding disease mechanisms. 
These experts are designed to capture distinct yet complementary pathological features. By the combination of the mentioned experts' knowledge, we obtain $E=\{E_C, E_T, E_O, E_B\}$ to analyze images, where  $E_C, E_T, E_O, E_B$ represent the expert-knowledge from the Cellular, Tissue, Organ, and Biomarker Experts, respectively. Each expert's knowledge is generated by the MLLM with the guidance $I_{G}$ in the decision information $I_{dec}$.

Inspired by Mixture of Experts (MoE) \cite{cai2024survey} that not all experts are required for image analysis, we select related experts based on specific questions (as discussed in Sec.~\ref{sec: Preparation}). Only the selected experts analyze the image, and their insights are aggregated and synthesized to form the final expert-knowledge $E$. 

\subsection{Summary \& Answer Generation}
\label{sec: summary}
Once the preparation and analysis stages are completed, PathCoT summarizes all acquired information including Image $I$, Question $Q$, image caption $D_{cap}$ and expert-knowledge $E$  as $S=\{I, Q, D_{cap}, E\}$. Finally, PathCoT generates the CoT-based answer $A_{CoT}$ as follows:
\begin{equation}
   A_{CoT} = MLLM(S).
\end{equation}

\subsection{Self-Evaluation}
\label{sec: Self-Evaluation}
To improve the reliability of results, PathCoT re-evaluates CoT-based answers since the additional reasoning steps in CoT may introduce errors.
These errors may cause the CoT-based answers to be less accurate than the direct reasoning answers, leading to the divergence of answers.
To address such divergence, PathCoT performs a self-evaluation stage that assesses the generated answers. In addition to the answer $A_{CoT}$ generated by CoT, we obtain another direct reasoning answer $A_{dir}$ generated by the MLLM without the CoT reasoning process:
\begin{equation}
   A_{dir} = MLLM(I, Q).
\end{equation}
Then, PathCoT considers both answers as candidates and determines the correct one (as shown in Fig.~\ref{fig: framework}).
When faced with conflicting candidate answers, PathCoT should provide the additional rationale $R$ for the decision, including the justification for selecting the correct answer and the explanation for rejecting the incorrect answer. Therefore, the final answer $A$ can be obtained by: 
\begin{equation}
    R, A = MLLM(I, Q, A_{CoT}, A_{dir}).
\end{equation}

\section{Experiments}

\begin{table*}[ht]
    \centering
    \caption{Performance comparisons of different methods on the PathMMU dataset.}
    \label{tab: comparisons}
    \resizebox{\linewidth}{!}{
        \begin{tabular}{c c c  c c  c c  c c  c c} 
        \hline
        \multirow{3}{*}{Method} & 
        \multicolumn{2}{c}{PubMed} & 
        \multicolumn{2}{c}{EduContent} & \multicolumn{2}{c}{Atlas} & 
        \multicolumn{2}{c}{PathCLS}& 
        \multicolumn{2}{c}{Overall}\\
         & Tiny Test & Test & Tiny Test & Test & Tiny Test & Test & Tiny Test & Test & Tiny Test & Test\\
         & (281) & (2787) & (255) & (1683) & (208) & (799) & (177) & (1632) & (921) & (6901) \\
        \hline
         MLLM Only & 
         39.15 & 36.35 & 
         33.33 & 35.59 & 
         37.02 & 34.79 & 
         21.47 &	19.98 & 
         33.66 & 32.11 \\
         MMCoT & 
         40.93 & 35.24 & 
         30.59 & 36.07 & 
         33.65 & 32.79 & 
         20.90 &	21.63 & 
         32.57 & 31.94 \\
         DDCoT & 
         31.67 & 31.40 & 
         28.63 & 32.20 & 
         31.73 & 31.54 & 
         18.08 &	17.77 & 
         28.23 & 28.39 \\
         Cantor & 
         38.08 & 34.09 & 
         36.86 & 33.21 & 
         37.50 & 34.92 & 
         18.64 &	18.32 & 
         33.88 & 30.24 \\         
         CCoT & 
         41.64 & 35.13 & 
         32.16 & 34.82 & 
         34.62 & 33.42 & 
         22.03 &	22.06 & 
         33.66 & 31.76 \\
         Qvix & 
         35.23 & 32.62 & 
         34.90 & 34.88 & 
         34.62 & 33.79 & 
         21.47 &	18.44 & 
         32.36 & 29.95 \\
         MC-CoT & 
         39.86 & 36.96 & 
         35.29 & 36.19 & 
         38.46 & 35.29 & 
         19.77 &	21.20 & 
         34.42 & 32.85 \\
         PathCoT &
         \textbf{45.55} & \textbf{40.90} & 
         \textbf{39.22} & \textbf{40.17} & 
         \textbf{42.79} & \textbf{42.43} & 
         \textbf{24.86} & \textbf{23.44} & 
         \textbf{39.20} & \textbf{36.75} \\
        \hline
        \end{tabular}
    }
    \end{table*}

\begin{table*}[ht]
    \centering
    \caption{Ablation studies for each stage in PathCoT.}
    \label{tab: ab_stage}
    \resizebox{\linewidth}{!}{
        \begin{tabular}{c c c  c c  c c  c c  c c} 
        \hline
        \multirow{2}{*}{Method} & 
        \multicolumn{2}{c}{PubMed} & 
        \multicolumn{2}{c}{EduContent} & \multicolumn{2}{c}{Atlas} & 
        \multicolumn{2}{c}{PathCLS}& 
        \multicolumn{2}{c}{Overall}\\
         & Tiny Test & Test & Tiny Test & Test & Tiny Test & Test & Tiny Test & Test & Tiny Test & Test\\
        \hline
         w/o Caption & 
         44.13 & 39.76 & 
         36.86 & 38.27 & 
         40.38 & 40.55 & 
         23.16 &	22.79 & 
         37.24 & 35.47 \\
         w/ Vanilla Caption & 
         44.84 & 40.04 & 
         37.65 & 39.28 & 
         41.35 & 41.43 & 
         23.73 &	22.98 & 
         38.00 & 35.98 \\
         w/o Analysis & 
         42.35 & 38.75 &
         34.51 & 36.84 & 
         38.46 & 38.92 & 
         22.60 &	21.38 & 
         35.50 & 34.20 \\
         w/o Self-Evaluation & 
         43.06 & 39.22 & 
         37.25 & 37.97 & 
         39.42 & 40.43 & 
         23.16 &	22.67 & 
         36.81 & 35.14 \\
         PathCoT &
         \textbf{45.55} & \textbf{40.90} & 
         \textbf{39.22} & \textbf{40.17} & 
         \textbf{42.79} & \textbf{42.43} & 
         \textbf{24.86} & \textbf{23.44} & 
         \textbf{39.20} & \textbf{36.75} \\
        \hline
        \end{tabular}
    }
    \end{table*}

\begin{table*}[ht]
    \centering
    \caption{Ablation studies for each expert in PathCoT.}
    \label{tab: ab_expert}
    \resizebox{\linewidth}{!}{
        \begin{tabular}{c c c  c c  c c  c c  c c} 
        \hline
        \multirow{2}{*}{Method} & 
        \multicolumn{2}{c}{PubMed} & 
        \multicolumn{2}{c}{EduContent} & \multicolumn{2}{c}{Atlas} & 
        \multicolumn{2}{c}{PathCLS}& 
        \multicolumn{2}{c}{Overall}\\
         & Tiny Test & Test & Tiny Test & Test & Tiny Test & Test & Tiny Test & Test & Tiny Test & Test\\
        \hline
         w/o Cellular Expert & 
         43.42 & 39.40 & 
         36.08 & 38.15 & 
         39.90 & 40.30 & 
         23.16 &	22.55 & 
         36.70 & 35.21 \\
         w/o Tissue Expert & 
         43.77 & 39.54 & 
         36.47 & 38.32 & 
         40.38 & 40.80 & 
         23.16 &	22.37 & 
         37.02 & 35.33 \\
         w/o Organ Expert & 
         44.84 & 40.11 & 
         37.65 & 39.22 & 
         41.38 & 41.80 & 
         24.29 &	23.10 & 
         38.11 & 36.07 \\
         w/o Biomarker Expert & 
         44.13 & 39.94 & 
         37.25 & 38.80 & 
         40.87 & 41.55 & 
         23.73 &	22.86 & 
         37.57 & 35.81 \\
         PathCoT &
         \textbf{45.55} & \textbf{40.90} & 
         \textbf{39.22} & \textbf{40.17} & 
         \textbf{42.79} & \textbf{42.43} & 
         \textbf{24.86} & \textbf{23.44} & 
         \textbf{39.20} & \textbf{36.75} \\
        \hline
        \end{tabular}
    }
    \end{table*}

\subsection{Dataset and Settings}
We evaluate the performance of PathCoT on the PathMMU dataset \cite{sun2024pathmmu}, a massive multimodal expert-level benchmark for understanding and reasoning in pathology. It includes 33,573 Q\&As along with 21,599 pathology images. 
We conduct our experiments on four subsets of PathMMU: PubMed, SocialPath, Atlas, and PathCLS. Each subset contains two test data: the tiny test set and test set, we evaluate our method on both of them. 

PathCoT is implemented using PyTorch \cite{paszke2019pytorch} on a single GPU (NVIDIA GeForce RTX 3090). We use LLava as the MLLM in PathCoT. For fair comparison, all comparison methods also utilize LLaVa as the MLLM. GPT-3.5 is employed as the language tool if an LLM is required.

\subsection{Baselines and Metric}
To demonstrate the effectiveness of PathCoT, we compare it with state-of-the-art (SOTA) zero-shot CoT-based methods as follows: 
\textbf{MLLM-Only}: directly applies the MLLM model to answer the question based on the image. 
\textbf{MMCoT} \cite{zhang2023multimodal}: generates rationales based on ground truth before producing the final answer.
\textbf{DDCoT} \cite{zheng2023ddcot}: uses the LLM to break down the problem into multiple related sub-questions and then answers each sub-questions.
\textbf{CCoT} \cite{mitra2024compositional}: obtains scene graphs and integrates it as the prompt for visual reasoning.
\textbf{Cantor} \cite{gao2024cantor}: utilizes the advanced cognitive abilities of the MLLM for general high-level logical problem-solving.
\textbf{Qvix} \cite{yang2023good}: applies MLLM to explore visual information based on the knowledge derived from LLMs.
\textbf{MC-CoT} \cite{wei2024mc}: designs a modular collaborative CoT framework for med-vqa tasks by integrating LLMs into the problem-solving process.
We evaluate our method by the index of Accuracy since all the questions in the test sets are in the form of multiple-choice questions.

\subsection{Comparisons with SOTA CoT-based methods}
In Tab.~\ref{tab: comparisons}, we compare our method with other zero-shot CoT-based methods on the PathMMU dataset. From Tab.~\ref{tab: comparisons}, we can draw the following observations: Firstly, these general CoT-based methods, which are not specifically designed for pathology images, perform poorly on pathology visual reasoning tasks due to the lack of domain-specific information, resulting in lower accuracy than results generated directly by MLLMs. Secondly, both MC-CoT and PathCoT introduce specific module to handle pathology images, and achieve better performance.
Finally, PathCoT consistently outperforms all comparison methods as it takes into account both expert-knowledge and divergence of answers during reasoning.

\subsection{Ablation Study}
\subsubsection{Impact of different stages.} We conduct the experiments to verify the effectiveness of different stages in PathCoT, and summarize the results in Tab~\ref{tab: ab_stage}. 
For ablating the Preparation stage, we establish two baselines: one without the caption (``w/o Caption'') and one with the vanilla caption (``w/ Vanilla Caption''). The ``w/o Caption'' baseline predicts the answers without generating captions, while the ``w/ Vanilla Caption'' baseline directly generates captions without considering the question-agnostic and question-dependent descriptions. From the first two rows of the Tab.~\ref{tab: ab_stage}, we can observe that both the vanilla caption and our designed caption can improve the reasoning accuracy. Furthermore, our designed caption can lead to an increase of 1.20\% and 0.77\% on the overall tiny test and test sets, respectively, compared to the vanilla caption baseline. These results demonstrate the importance of question-agnostic and question-dependent descriptions in helping MLLMs better understand the image and answer the given questions more accurately. 
In addition, the analysis stage improves the results by 3.70\% and 2.55\% on the overall tiny test and test sets, respectively, as shown in the third row of Tab.~\ref{tab: ab_stage}. This improvement is driven by the integration of knowledge from different experts in analyzing pathology images, which benefits visual reasoning tasks.
Finally, results are improved by 2.39\% and 1.61\% with the self-evaluation stage (fourth row of Tab.~\ref{tab: ab_stage}). This demonstrates that the proposed self-evaluation process can effectively address divergence of answers and improves the overall accuracy of the results. 

\subsubsection{Impact of different experts.}
To highlight the importance of each expert in PathCoT, we present the results in Tab.~\ref{tab: ab_expert} when specific expert is excluded. The results indicate that each expert contributes positively to PathCoT. Moreover, we observe that the Cellular and Tissue experts are more effective than other experts, implying their crucial role in pathology visual reasoning tasks.

\begin{figure}[t]
  \centering
  \includegraphics[width=1.\linewidth]{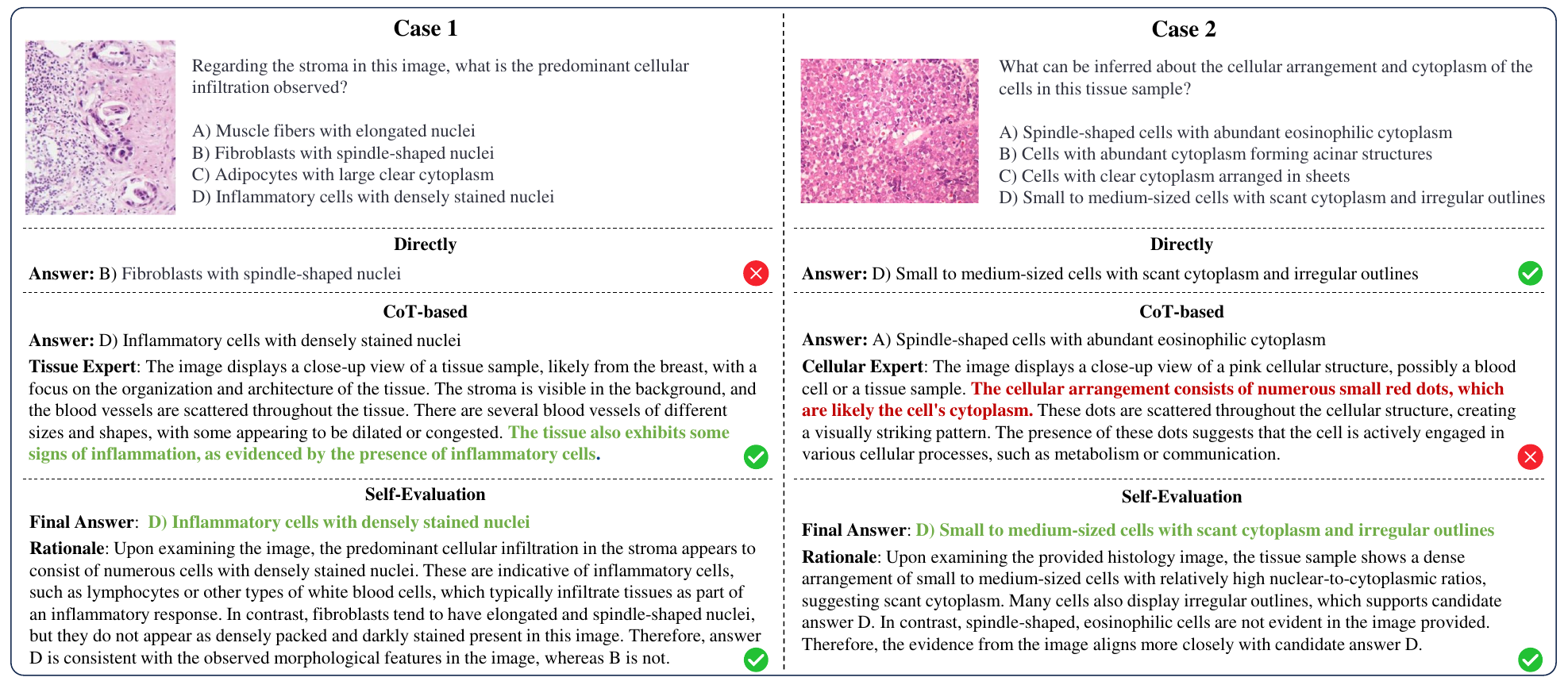}
   \caption{The cases of PathCoT.}
   \label{fig: cases}
\end{figure}

\subsection{Case Presentation}
In this section, we provide two cases to demonstrate how PathCoT effectively utilizes reasoning chains to derive correct results, as shown in Fig.~\ref{fig: cases}.
 
In Case 1, we present how expert-knowledge can significantly aid the reasoning process by providing valuable information. For instance, the Tissue Expert provides the knowledge: \emph{The tissue also exhibits some signs of inflammation, as evidenced by the presence of inflammatory cells}, which helps the MLLM perform reliable reasoning and obtain the right answer.

In Case 2, we show how self-evaluation addresses the divergence of answers. As shown in Fig.~\ref{fig: cases}, the knowledge from the Cellular Expert: \emph{The cellular arrangement consists of numerous small red dots, which are likely the cell's cytoplasm}, introduces error information, and finally results in incorrect answer. However, MLLM is able to predict the right answer by direct reasoning. In face with the divergent answers, PathCoT performs the self-evaluation that can choose the right answer and provide rationales for the decision. 

\section{Conclusion}
We propose PathCoT, a novel CoT prompting method for zero-shot pathology visual reasoning, which integrates the pathology expert-knowledge into the reasoning process of MLLMs and incorporates self-evaluation to mitigate divergence of answers in CoT reasoning. The experimental results  verify the efficacy of our method on pathology visual understanding and reasoning.


%
%
%
%

\bibliographystyle{unsrt}
\bibliography{ref}

\end{document}